\def\year{2022}\relax
\documentclass[letterpaper]{article} 
\usepackage{aaai22}  
\usepackage{times}  
\usepackage{helvet}  
\usepackage{courier}  
\usepackage[hyphens]{url}  
\usepackage{graphicx} 
\usepackage{natbib}  
\usepackage{caption} 
\DeclareCaptionStyle{ruled}{labelfont=normalfont,labelsep=colon,strut=off} 
\usepackage{algorithm}
\usepackage{algorithmic}

\usepackage{newfloat}
\usepackage{listings}
\floatstyle{ruled}
\newfloat{listing}{tb}{lst}{}
\floatname{listing}{Listing}
\usepackage{array}
\usepackage{xcolor}

\newcommand{\modelname}{SC-AAE}
\newcommand{\fsname}{FS-GER}
\newcommand{\accuracy}{$58.43\%$ }
\usepackage{multirow, makecell}
\usepackage{subcaption}
\usepackage{booktabs}
\usepackage{amsmath}
\usepackage{amssymb}
\usepackage{amsfonts}
\usepackage{enumitem}
\usepackage{float}
\usepackage{bbm}
\usepackage{dsfont}
\usepackage{bbold} 
\usepackage{hhline}

\newcommand{\shorteq}{%
  \settowidth{\@tempdima}{-}
  \resizebox{\@tempdima}{\height}{=}%
}

\newcolumntype{L}[1]{>{\raggedright\let\newline\\\arraybackslash\hspace{0pt}}m{#1}}
\newcolumntype{C}[1]{>{\centering\let\newline\\\arraybackslash\hspace{0pt}}m{#1}}
\newcolumntype{R}[1]{>{\raggedleft\let\newline\\\arraybackslash\hspace{0pt}}m{#1}}

\title{Learning Unseen Emotions from Gestures via Semantically-Conditioned Zero-Shot Perception with Adversarial Autoencoders}
\author{Abhishek Banerjee, Uttaran Bhattacharya, Aniket Bera\\ 
\footnotesize{Department of Computer Science, University of Maryland}\\ 
College Park, Maryland 20740, USA\\
\{abanerj8, uttaranb, bera\}@umd.edu\\ 
\texttt{https://youtu.be/yalthVwTJ5s}
}

\begin{document}

\maketitle

\begin{abstract}
We present a novel generalized zero-shot algorithm to recognize perceived emotions from gestures. Our task is to map gestures to novel emotion categories not encountered in training. We introduce an adversarial autoencoder-based representation learning that correlates 3D motion-captured gesture sequences with the vectorized representation of the natural-language perceived emotion terms using \textit{word2vec} embeddings. The language-semantic embedding provides a representation of the emotion label space, and we leverage this underlying distribution to map the gesture sequences to the appropriate categorical emotion labels. We train our method using a combination of gestures annotated with known emotion terms and gestures not annotated with any emotions. We evaluate our method on the MPI Emotional Body Expressions Database (EBEDB) and obtain an accuracy of \accuracy. We see an improvement in performance compared to current state-of-the-art algorithms for generalized zero-shot learning by $25$--$27\%$ on the absolute. We also demonstrate our approach on publicly available videos from the internet and movie scenes, where the actors' pose has been extracted and map to their respective emotive states.
\end{abstract}

\section{Introduction}

\noindent \begin{figure}[t]
    \centering
    \includegraphics[width=\columnwidth]{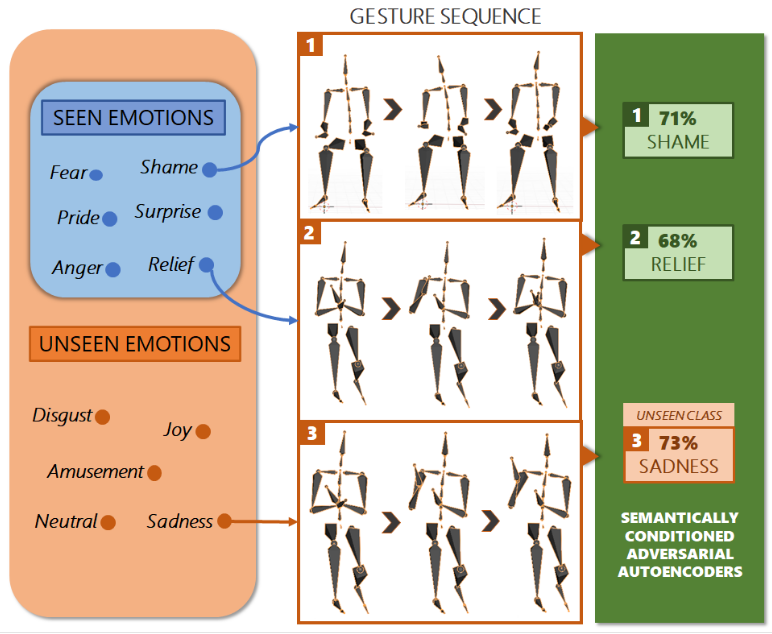}
    \caption{\textit{\textbf{Generalized zero-shot Emotion Recognition from gestures.} Gesture sequences from both seen and unseen classes of emotions are used as the input to our AAE-based representation learning algorithm. We capture the spatial-temporal representation of 3D motion-captured gesture sequences in our network and correlate them with the semantic representation of the corresponding perceived emotion term. Our network can accurately recognize emotions not seen during training and has an overall accuracy of \accuracy.}}
    \label{fig:intro}
    \vspace{-10pt}
\end{figure}

\begin{figure}[h]
    \centering
    \includegraphics[width=\columnwidth]{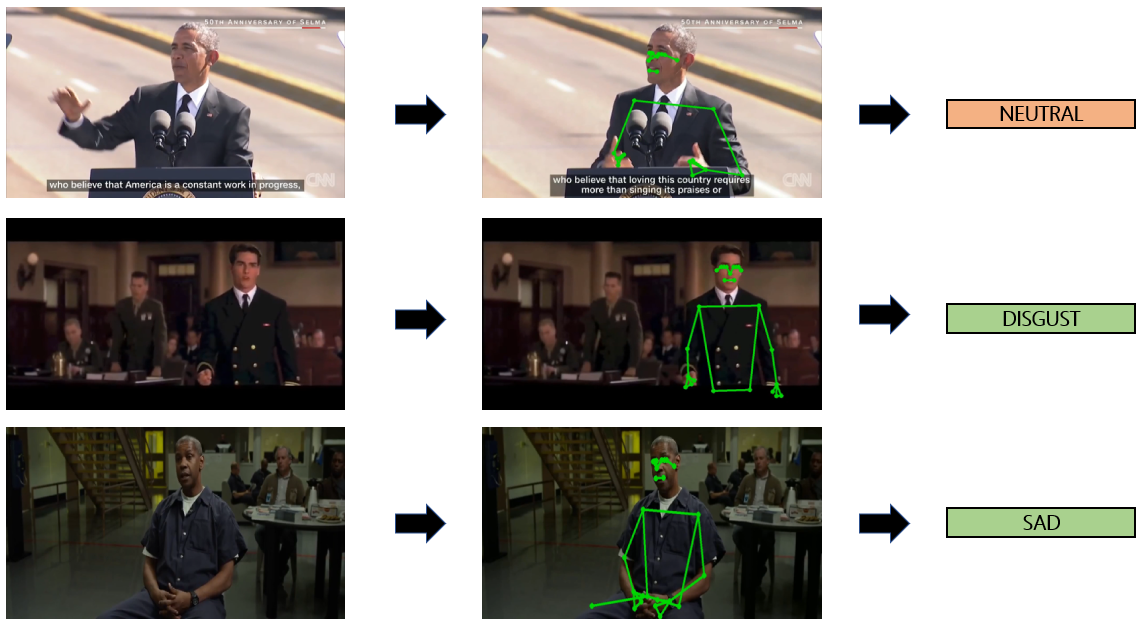}
    \caption{\textit{\textbf{Public Video Results}. In order to validate our results, we use publicly available videos wherein the individuals are unambiguous in their emotive state. Using the network introduced in \cite{pavllo20193d}, we extract the 3D pose of the person and then feed that into our network to recognize the emotion.}}
    \label{fig:real-world}
\end{figure}

Emotion recognition as an area of research is integral to a variety of domains, including human-computer interaction, robotics~\cite{liu2017facial} and affective computing~\cite{yates2017arousal}. Existing research in emotion recognition has leveraged aspects such as facial expressions~\cite{liu2017facial}, speech and gaits~\cite{bhattacharya2020step} to gauge an individual's emotional state. Gestures have also been used in psychological studies to identify emotions by using affective features such as arm swing rate, posture, frequency of movements, etc. Recent work by~\cite{bhattacharya2020step} has leveraged spatial-temporal graph convolution networks (ST-GCN)~\cite{yan2018spatial} to capture pose dynamics and generate a mapping between the extracted features and the labeled emotions.

A major challenge in machine learning-based emotion recognition algorithms is the requirement for significantly-sized, well-labeled datasets to build classification algorithms on previously labeled emotions. However, considering the wide spectrum of emotions for humans~\cite{zhou2016emotion} and different emotion representations, it is tedious and often prohibitively expensive to develop datasets with an adequate number of instances for every emotion. Zero-shot learning has recently drawn considerable attention to overcome such issues where labels of different classes are unavailable. It provides an alternative methodology that does not rely on existing labels. Instead, it relies on utilizing the different relationships between various seen and unseen classes to determine the appropriate labels.

\begin{figure*}[t]
    \centering
    \includegraphics[width=0.92\textwidth]{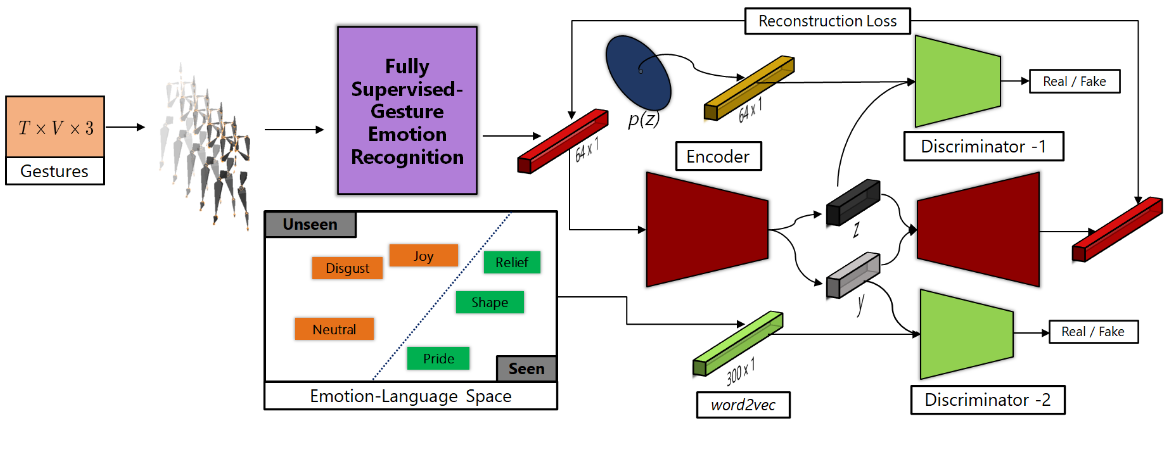}
    \caption{\textit{\textbf{Network overview.} Our network consists of a feature extraction pipeline that takes the sequence of gestures ($T$: time steps, $V$: joints or nodes) and extracts the relevant high-level features. These features are fed into a semantically-conditioned adversarial autoencoder, which projects these features onto a latent space by aligning it with the word-level semantic information from the \textit{word2vec} dimension. The network's encoder generates two latent vectors corresponding to the embeddings for the gestures and the word embeddings. The adversarial loss is used by the two discriminators to align the latent distributions to the corresponding priors.}}
    \label{fig:pipeline}
    \vspace{-12pt}
\end{figure*}

In the generalized zero-shot learning (GZSL) paradigm, a network learns to recognize all classes, seen and unseen, while being trained with data annotations available only for seen classes. The model learns to generalize on the unseen classes by leveraging information from other modalities, such as language semantics, to create class embeddings corresponding to each label. Recent approaches to the zero-shot problem have used generative models~\cite{mishra2018generative} to synthesize features for the unseen classes, which are then used for the classification task. GANs and VAEs have been the most prominent methods to synthesize these features; however, Shi et al.~\cite{shi2019variational} have shown that the representation of multi-modal distributions by VAEs can result in sub-optimally learned representations. While GANs can create higher quality features than VAEs, the latent distribution spaces they learn can be susceptible to mode collapse\cite{goodfellow2014generative}. 

On the other hand, adversarial autoencoders (AAEs) create more closely aligned latent distributions than VAEs or GANs~\cite{makhzani2015adversarial}. Therefore, we build on the network by Makhazani et al.~\cite{makhzani2015adversarial} to develop our network architecture.

\noindent{\bf Main Results:} We present a generalized zero-shot algorithm to recognize perceived emotions from 3D motion-captured gesture sequences represented as upper-body poses. To capture the semantic relationships between the emotion classes, we leverage the rich word embeddings of the pre-trained \textit{word2vec} model~\cite{mikolov2013efficient}. A fully-supervised emotion recognition network generates a feature vector corresponding to a sequence of gesture inputs. We use an autoencoder architecture coupled with an adversarial loss to generate latent representations for the learned gesture-feature vectors learned from the fully supervised network
corresponding to gesture sequences and another adversarial loss to align these latent representations with the semantically conditioned distribution space of the emotion classes. Our main contributions include:
\begin{enumerate}
    \item A GZSL algorithm, {\tt \modelname}, based on an adversarial autoencoder architecture. We train it to learn a mapping between the gesture-feature vectors corresponding to 3D motion-captured gesture sequences and the seen and unseen perceived emotion classes expressed in natural language. To the best of our knowledge, our method is among the first to classify unseen perceptual affective labels in a zero-shot learning fashion.
    \item A fully supervised emotion recognition algorithm, {\tt \fsname} that classifies 3D motion-captured gesture sequences seen emotion classes. We use this architecture to generate the feature vectors for input to our {\tt \modelname} for GZSL.
\end{enumerate}

Our fully supervised network achieves a validation accuracy of $77.61\%$ with the seen emotion classes in the MPI Emotional Body Expressions Database (EBEDB)~\cite{volkova2014emotion}, which outperforms state-of-the-art methods for fully supervised action and emotion recognition by $7$--$18\%$ on the absolute. More importantly, we achieve an accuracy of \accuracy on EBEDB over the collective set of 11 seen and unseen emotion classes, outperforming state-of-the-art ZSL methods by $25$--$27\%$ on the absolute.

\section{Related Work}
We provide an overview of emotion representation, emotion recognition from non-verbal body expressions, and relevant developments in Zero-Shot learning.

\subsection{Emotion Recognition}
Recent works~\cite{sanders2016gait} in emotion recognition showcase the correlation between gaits and inherent psychological stress. Methods of~\cite{sapinski2019emotion} use deep learning methods to identify emotion states from gestures extracted from videos. Studies by~\cite{wegrzyn2017mapping} identified people's emotional states through psychological studies of human facial expressions. With the advent of deep learning, various works have emerged that use vision-based methods~\cite{akputu2013facial} to determine emotional state from facial expressions or audio signals using speech~\cite{deng2017semisupervised}. Recently, a number of works have used multiple modalities, including speech and facial expressions, in determining emotions~\cite{albanie2018emotion}.

\subsection{Generalized Zero-Shot Learning}
\label{sec:generative}
In the Generalized Zero-Shot Learning (GSZL) problem, the recognition task is executed for both seen and unseen classes. In contrast, for Zero-Shot Learning (ZSL), recognition is attempted on only the unseen classes alone. The GZSL is more challenging than the nominal-ZSL, where the model classifies only unseen classes because of the hubness problem~\cite{dinu2014improving}, which occurs when the model overfits to the trained classes.
Recently, generative methods have become popular in GZSL, which uses either GANs~\cite{mishra2018generative} or VAEs~\cite{schonfeld2019generalized} to generate features for unseen classes. Traditional GZSL generative models rely on a data augmentation method, which generates features of interest that have been hitherto unseen by the model during training. 

Hubert et al.~\cite{hubert2017learning} have shown that mapping the joint visual-language features to a latent space instead of the language space gives higher accuracy. ~\cite{schonfeld2019generalized} use unconditional VAEs and achieve multi-modal alignment via cross-reconstruction and distribution alignment. In our algorithm, we build on the network model used by~\cite{schonfeld2019generalized} to perform our latent space embedding and classification task. Considering the multiple modalities that are used to learn the distribution in our approach, \textit{i.e.}, language semantics for emotions and gestures, we rely on methods that correlate the learned distributions of these modalities to estimate the semantic relation between the classes accurately.

\begin{figure*}[t]
    \centering
    \includegraphics[width=\linewidth]{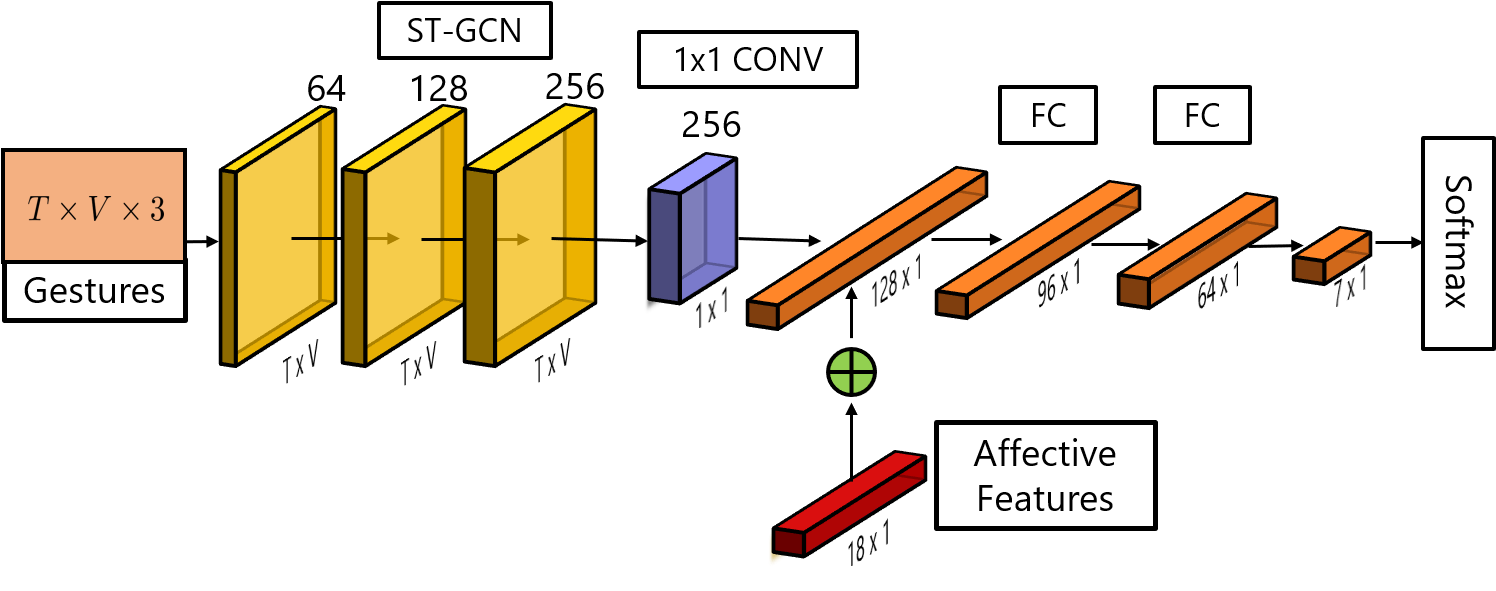}
    \caption{\textit{\textbf{Fully Supervised Network for Emotion Recognition from Gestures (\fsname).} The network comprises of three ST-GCN layers, followed by a single 1$\times$1 convolution layer. The input data is of the form $T$: time Steps (510 at 30fps) $\times$ $V$: nodes (10 joints) $\times$ 3 (dimension of nodes). The convolution output is appended with the affective features, $A$, and then passed through subsequent Fully Connected (FC) layers to generate a 64-dimensional feature description vector. This layer is passed through an FC of size 7 (total number of classes on which it is trained). The softmax layer uses this for classification. The 64-dimensional embedding is extracted from the network after the fully supervised step for the GZSL task.}}
    \label{fig:STEP}
    \vspace{-12pt}
\end{figure*}

\section{Method}
In this section, we define the problem statement and describe our approach in detail. We present an overview of our proposed algorithm in Figure~\ref{fig:pipeline}. We use the sequence of 3D motion-captured poses as input to our pre-trained feature extraction module to obtain the corresponding feature vectors. Subsequently, using the \textit{word2vec} representation, we obtain the semantic word-level embedding for the specific emotion. The semantic embedding and the corresponding feature vector are used as inputs to the {\tt \modelname} architecture. The encoder part of the VAE outputs a class semantic label as well as a latent vector. The two vectors are passed through two subsequent discriminators that use the adversarial loss to increase the encoder's estimated classification accuracy. For classification, we use the encoder to output the corresponding semantic labels, which are then matched with the relevant class labels.

\subsection{Problem Definition}
We discuss the task definition of our approach in this section. Let
$S=\{(x, y, c(y)) \mid x \in X, y \in Y^S, c(y) \in C$ be a set of input data. Here $x$ denotes an input vector embedding representing a sequence of gestures, and $y$ is the corresponding class label, which in our approach is the associated emotion. $c(y)$ is the semantic embedding corresponding to the class label. In our work, we use the \textit{word2vec} representation for the semantic description (described later in Section~\ref{section:textem}). We also have the auxiliary training set $U=\left\{(u, c(u)) \mid u \in Y^{U}, c(u) \in C\right\}$, for all the unseen classes. Here, $u$ denotes an unseen class from the set $Y^U$, which is disjoint from $Y^S$. The task that we have at hand is the GZSL task, which evaluates the network on both seen and unseen classes, denoted by $f_{GZSL} : X \rightarrow Y^S \cup Y^U$.

We approach our problem of GZSL in the transductive setting~\cite{wan2019transductive}.
In the transductive setting, the pre-trained network has access to the unseen classes, but the data points in these classes do not have any associated labels. We create a single dummy label for all the gestures belonging to all the unseen classes during feature generation using the pre-trained network.
We first start with a description of the feature generation module, which generates an embedding corresponding to the input sequence of gestures.

\subsection{Feature Extractor Network}
We show an overview of the feature extraction network in Figure~\ref{fig:STEP}. The input to the network is a sequence of poses of size $T$ (time steps) $\times$ $V$ (nodes) $\times$ 3 (position coordinates).
 Because gestures are a periodic sequence of poses, we use ST-GCN~\cite{yan2018spatial}, which captures spatial and temporal features of interest for the input gaits. The relationship between specific joints is defined using an adjacency matrix, and this relationship is leveraged while using the GCNs. The first ST-GCN layer has $64$-layers while the second and third have $128$ and $256$-layers, respectively. ST-GCN layers are followed by a ReLU activation function and a BatchNorm layer. The output of the convolution operation is passed through a $1 \times 1$ convolution layer, giving a $128$-dimensional vector. This is appended with an affective feature vector extracted from the gestures during pre-processing and is subsequently passed through two successive fully connected (FC) layers to give a $64 \times 1$ feature vector. This feature vector is passed through a fully connected layer, followed by a softmax layer to generate labels for classification. We treat the feature vectors belonging to unseen classes, $Y^{U}$, as a single class with a dummy label.

The gestures are predominantly in the upper part of the body; therefore, we consider only the relevant joints in the upper body. Affective features from gestures have been shown to be relevant to the problem of emotion recognition~\cite{randhavane2019identifying}, and consist of posture and motion features:

\setlist{nolistsep}
    \begin{itemize}[noitemsep]
    \item \textbf{\textit{Posture features}}: These consist of distances between pairs of joints, as well as angles and areas formed by three joints of interest.
    \item \textbf{\textit{Motion features}}: These consist of velocity and acceleration of joints of interest in the gesture.
\end{itemize}


Based on visual perception and psychology literature~\protect \cite{crenn2016body}, we use $18$ extracted features,  which we append to the output of the ST-GCN layers.


\subsection{Language Embedding}
\label{section:textem}
The key idea in our zero-shot learning is to utilize the semantic relationship between multiple classes of emotions to determine the association between various gesture sequences and the seen and unseen emotion classes. The \textit{word2vec}~\cite{mikolov2013efficient} representation gives a $300$-dimensional embedding vector based on the semantics of the word. Using the vector representations for all emotions, we can ascertain the level of ``closeness" or ``disparity" between them. For the unseen classes, these representations give us the underlying relationship between instances of that class and other classes in the seen and unseen domains, allowing us to classify them into the appropriate categories.

We represent the set of emotions as
\begin{equation}
    \mathcal{E}=\{e_1,e_2,e_3,....,e_n\},
    \label{eqn:word2vec}
\end{equation}
where $\{e_i\} \in \mathbb{R}^{300}$ is the \textit{word2vec} representation of the emotion-word. This way, two specific emotions can be related by Euclidean $\ell_2$-norm distance to ascertain their adjacency.

\subsection{Adversarial Autoencoder}
In our current method, we build on the work of~\cite{makhzani2015adversarial} to create an adversarial autoencoder, which learns from the semantic distributions of data in the language space as well as the gesture space. The VAE in such a setting is regularized by matching the posterior $q(z\mid x)$ to a prior $p(z)$ distribution. The training of the network takes place in two phases:
\begin{itemize}
    \item the \textit{reconstruction phase}, where the autoencoder updates the encoder and the decoder to minimize the reconstruction error of the inputs, and
    \item the \textit{regularization phase}, where the adversarial network first updates its discriminative network to separate the true samples from the generated samples. The generator we use to compute the adversarial loss in our case comes from the encoder network of the VAE.
\end{itemize}

\subsection{Network Architecture}
As seen in Figure~\ref{fig:pipeline}, {\tt \fsname} outputs a $64$-dimension feature vector for the respective gesture input sequence. Correspondingly, we get the $300$-dimension language embedding using \textit{word2vec}. The encoder for the adversarial autoencoder (AAE) predicts the latent vector corresponding to the gesture $z$ and the class semantic label, $\hat{y}$. The generated labels and vectors are then passed through two separate discriminators that help discriminate between the desired samples from the prior and those generated by the encoder. After the training, we use the encoder to generate the relevant semantic labels, which identifies the predicted emotion label corresponding to that gesture-sequence input.

\subsection{Loss Functions}
\label{section:loss}
We aim to minimize the cross-alignment loss between gestures and the word-labels. As we have two separate modalities, we utilize two separate VAEs akin to those in~\cite{schonfeld2019generalized} to map the inputs to a common latent space. We can write the loss as
\begin{equation}
    \resizebox{\columnwidth}{!}{%
        $\mathcal{L}_{VAE}= \mathbb{E}_{q_{\phi}(z \mid x)}\left[\log p_{\theta}\left(x^{(i)} \mid z\right)\right]-\beta D_{K L}\left(q_{\phi}\left(z \mid x^{(i)}\right) \| p_{\theta}(z)\right)$.
    }
\end{equation}
The KL divergence aligns the desired distributions. In our algorithm, we use the adversarial losses to align the prior distributions with the encoder output; hence we do not use the KL divergence.

\subsubsection{Adversarial Loss}
We can write the adversarial loss for a discriminator as
\begin{equation}
    \resizebox{\columnwidth}{!}{%
        $\mathcal{L}_{ADV} = \mathbb{E}_{\mathbf{x} \sim p(\mathbf{x})}\left[\log D(\mathbf{x})\right] +\mathbb{E}_{\overline{\mathbf{x}} \sim p_{\theta}(\tilde{\mathbf{x}} \mid \mathbf{z}, \mathbf{a})}\left[\log \left(1-D(\tilde{\mathbf{x}})\right)\right]$
    }
\end{equation}
There are two adversarial losses used in our network, corresponding to two discriminators. For the label discriminator, $a$ corresponds to $c(y)$, which is an element of $\mathcal{E}$, in Equation~\ref{eqn:word2vec}. We denoted this by $\mathcal{L}_{ADV-lang}$. For the feature discriminator, it corresponds to an element from the generated features from a prior distribution $p(z)$ and we denote the adversarial loss for this by $\mathcal{L}_{ADV-feat}$

Hence, we can write our net loss as
\begin{equation}
\mathcal{L}_{NET}=\mathcal{L}_{VAE}+\gamma \mathcal{L}_{ADV-lang}+\delta \mathcal{L}_{ADV-feat},
\end{equation}
where $\gamma$ and $\delta$ are weighing functions.

\section{Results and Experiments}
We present experiments and results for our zero-shot classification task in this section, including the details of our network and the hardware configuration.

\subsection{Dataset}
We train and evaluate our network on the MPI Emotional Body Expressions Database (EBEDB)~\cite{volkova2014emotion}. It consists of $1,447$ 3D motion-captured sequences of natural-emotion body gestures from actors as they narrated specific lines. All body movements were captured at $120$ fps. The original dataset consists of information regarding $23$ joints in the body. However, because we are interested in gestures made by the upper body, we select $V=10$ joints: the head, neck, right-shoulder, left-shoulder, right-elbow, left-elbow, right-wrist, left-wrist, backbone, and pelvis. We ignore the lower-body joints as there is no significant motion in those joints. Each sequence is annotated with one of $11$ categorical emotion classes.

To evaluate our model, we split the $11$ available emotion classes in MPI EBEDB into a roughly equal split of six seen classes and five unseen classes. During the training phase, the model learns only from the six seen classes. Since there are multiple possible combinations for choosing these five unseen classes and there are no fixed criteria in particular for this dataset for zero-shot learning, we conduct five experiments in which we successively select five random classes from the available $11$. Our results are averaged over these five experiments. We use a train-test split of 80\%-20\%.

Currently, this is the only database that is publicly available that maps motion sequences of humans to their emotional states. On account of this, our discussions and evaluation take place only on this particular dataset. However, to validate our results, we use popular videos which are unambiguous with regards to the emotion state of the actors. Some of these are given in Figure \ref{fig:real-world}.

\subsection{Training Details}
\label{subsec:TrainingParams}
All our encoders and decoders are multi-layer perceptrons with two hidden layers. More hidden layers reduce the performance because the gesture-features and language embeddings are very high-level representations and generally sparse; hence more layers would result in loss of crucial features for classification. 

We use $100$ hidden units each for the encoder and the decoder. The discriminators consist of two hidden layers with $100$ hidden layers each for the language-embedding model, while the discriminator for the gesture-feature vector has two hidden layers of size $100$ and $32$, respectively. In our work, we use the proposed {\tt \fsname} to generate a $64$-dimension feature vector corresponding to the gestures and a $300$-dimension \textit{word2vec} feature encoding the emotions.

We train the model for $200$ epochs by stochastic gradient descent using the Adam optimizer~\cite{adam} and a batch size of $6$ for features. Each batch consists of pairs of extracted gesture features and matching attributes from different seen classes. Pairs of data always belong to the same class. We keep the values of $\gamma$ and $\delta$ constant and discuss how we choose their values in Section~\ref{subsubsec:hyerparameters}. Our network takes around 6 minutes to train on an Nvidia RTX 2080 GPU.

\begin{figure*}[t]
    \centering
    \includegraphics[width=\linewidth]{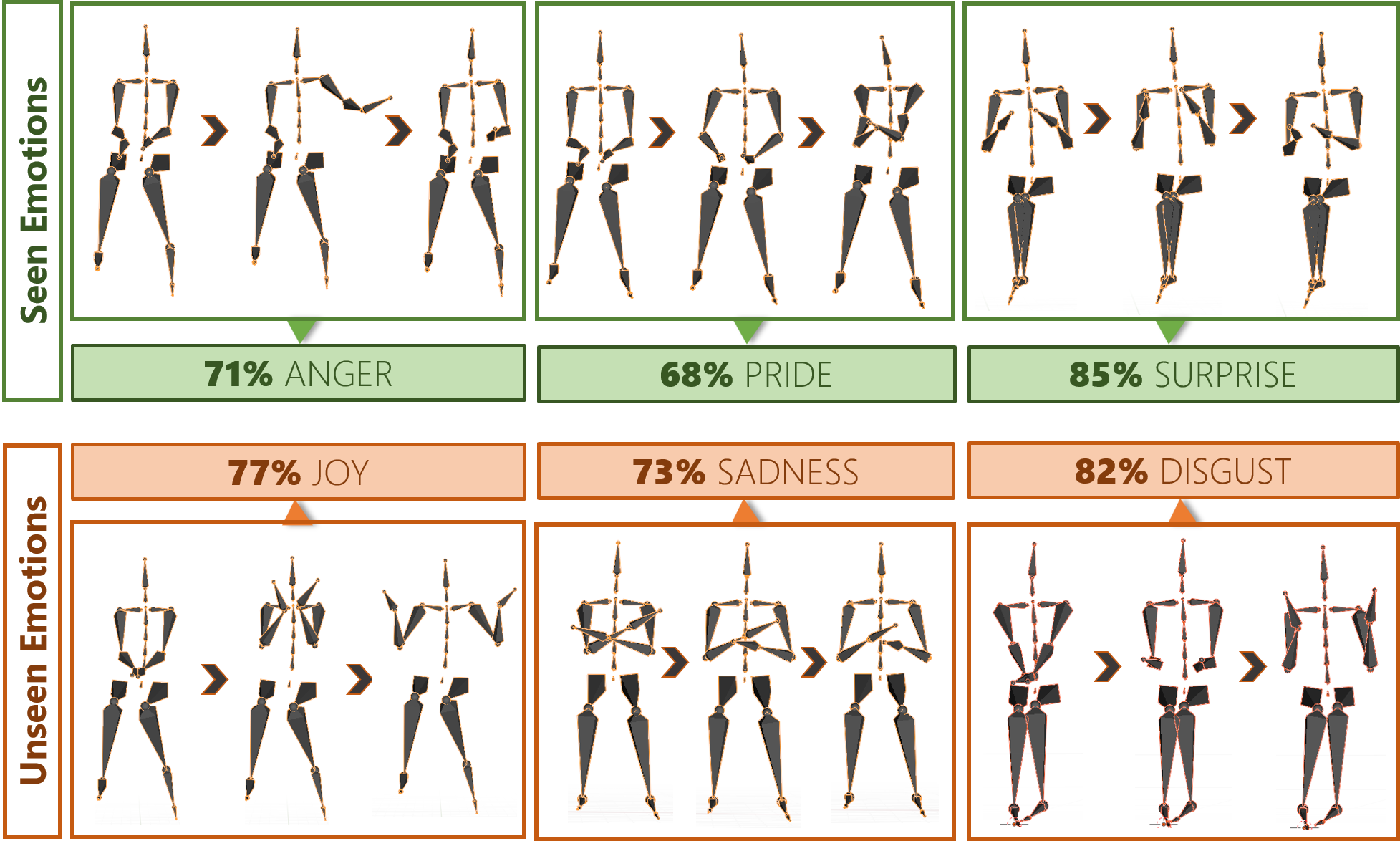}
    \caption{\textit{\textbf{Visual Results.} The top row shows three sets of gestures in temporal order from left-to-right, which map to the correct seen emotions during classification. The bottom row consists of three gestures mapped to the correct unseen emotions during training.}}
    \label{fig:Training}
    \vspace{-12pt}
\end{figure*}

\subsection{Performance of \fsname}
We compare the performance of {\tt \fsname} (Fully-Supervised Emotion Recognition) with previous methods on emotion recognition~\cite{bhattacharya2020step}, as well as action recognition~\cite{yan2018spatial}. In \cite{yan2018spatial}, the authors introduce ST-GCNs to perform action recognition. The network takes a sequence of gaits as input and uses the spatial relation between the various joints and their temporal locations to create a mapping between the motion sequences and their actions. In~\cite{bhattacharya2020step}, the authors develop an emotion-specific embedding method to augment the graph convolution network's ability to map motion patterns to perceived emotions. In addition to capturing the spatial and temporal variance of the joints, they extract certain affective features that capture semantics more specific to emotions. 

We show the overall network architecture for our network, {\tt \fsname} in Fig.~\ref{fig:STEP}.
We train all networks from scratch using all the body joints as per their input requirements. We classify for the same set of six seen classes and one dummy class corresponding to the five unseen classes. Based on the MPI EBEDB~\cite{volkova2014emotion}, we have $T= 510$ time steps and $V= 10$ joints in the upper body.

We report the performance of all the methods in Table~\ref{tab:fs_accuracy}. We observe that our method outperforms the other methods by $7$--$18$\% on the absolute, as a result of using the relevant set of joints and affective features. We use our proposed emotion classifier network to generate features for the subsequent GZSL framework.

\begin{table}[h]
    \centering
    \caption{Classification accuracies for fully-Supervised emotion recognition methods on the seen emotion classes (\textit{colored row is best}).}
    \label{tab:fs_accuracy}
    \begin{tabular}{lc}
        \toprule
        \textit{\textbf{Method}} & \textit{\textbf{Accuracy}} \\
        \midrule
        \textit{ST-GCN~\cite{xian2018feature}} & $59.12\%$ \\
        \textit{STEP~\cite{bhattacharya2020step}} & $70.38\%$ \\
        \textit{\tt \textbf{\fsname} \textbf{(Ours)}} & \textbf{$77.61\%$} \\
        \bottomrule
    \end{tabular}
    \vspace{-12pt}
\end{table}


\subsection{Related Methods}
We compare with ZSL methods for image classification and action recognition. Similar to our method, these methods also attempt to learn mappings from visual as well as spatial-temporal feature vectors to semantic descriptions.

\begin{itemize}
    
    \item \textbf{Image Classification.} We compare with state-of-the-art image classification problems in the GZSL paradigm, such as CADA-VAE~\cite{schonfeld2019generalized}, f-CLSWGAN~\cite{xian2018feature}, and CVAE-ZSL~\cite{mishra2018generative}. In ~\cite{schonfeld2019generalized}, the authors implement two separate VAEs use cross-reconstruction losses to align them. In~\cite{xian2018feature}, the authors use a GAN-based reconstruction to generate unseen features and leverage the Wasserstein distance to align the multiple-distributions. In~\cite{mishra2018generative}, the authors implement a standard VAE architecture and add semantic labels to the inputs for calculating the reconstruction loss.
\end{itemize}

For a fair comparison, we trained all these methods from scratch on MPI EBEDB~\cite{volkova2014emotion}.

\subsection{Evaluation Metric}
Following previous works in the GZSL paradigm  ~\cite{schonfeld2019generalized,xian2018feature}, we evaluate performance using the harmonic mean of the accuracies on the seen and the unseen classes. The harmonic mean is given by
\begin{equation}
    H=2 *\left(a c c_{\mathcal{Y}^{t r} *} a c c_{\mathcal{Y}^{t s}}\right) /\left(a c c_{\mathcal{Y}^{t r}}+a c c_{\mathcal{Y}^{t s}}\right),
\end{equation}
where $a c c y^{t r}$ and $a c c y^{t s}$ represent the accuracy of gestures from seen and unseen classes, respectively. The harmonic mean is preferred over the more conventional arithmetic mean in this paradigm because the arithmetic mean gives a large value if the seen class accuracy is much greater than the unseen class accuracy. By contrast, the harmonic mean only gives a large value both the seen and the unseen class accuracies are large, providing a more accurate reflection of performance.




\subsection{Evaluation of our Zero-Shot Framework}
We evaluate our proposed ZSL approach ({\tt \modelname}) with the other approaches for the GZSL task in Table~\ref{tab:accuracy_results}. We report the harmonic mean of the accuracies for the seen and the unseen classes, as achieved by each method. We observe that our proposed approach, {\tt \modelname}, outperforms the other approaches by $25$--$27$\% on the absolute. f-CLSWGAN~\cite{xian2018feature}, which conditioned GANs on image classification, suffers from mode collapse. CADA-VAE~\cite{schonfeld2019generalized}, while aligning the language-semantic and gesture-feature spaces effectively, fails to create representative features for the unseen classes, which can help in recognition. CVAE-ZSL ~\cite{mishra2018generative}, which was built for the action recognition task, does not generate robust features for emotion recognition. The visual results for our method can be seen in Figure \ref{fig:Training}.

\subsection{Analysis of the Zero-Shot Model}
In this section, we present an analysis of our zero-shot learning architecture, including the choice of hyperparameters and the size of the latent space. For additional analysis and details, please refer to the technical appendix.

\subsubsection{Hyperparameters}\label{subsubsec:hyerparameters}
Our model uses two hyperparameters, $\gamma$, and $\delta$, for regularizing the loss function for the network. These weigh the effect of the adversarial loss from both discriminators, \textit{i.e.}, from the language embedding and the extracted gait features, on the training process. Fixing $\gamma$ at $1$, we varied $\delta$ between $0.1$ and $2$ during training. On account of the heavier usage of the \textit{word2vec} embedding in the determination of classification accuracy, we found $\delta=1.5$ to give us the highest harmonic mean of accuracies, and therefore we have used this value to report our results. Changing $\gamma$ while keeping $\delta$ fixed at $1.5$ did not result in any significant changes, as these changes were largely overshadowed by the gains from changing $\delta$. Hence, we set $\gamma=1$ for our experiments.

\subsubsection{Size of Latent Embedding}
The latent embedding refers to the size of the gesture feature vector used in our latent space. We changed the sizes of the latent embeddings, $d$, from $d=2$ to $d=32$ in steps of one. We obtained the best results for $d=16$ and used this in our final network.

\begin{table}[t]
    \centering
    \caption{Harmonic mean of classification accuracies on seen and unseen classes by different methods on our GZSL task (\textit{colored row is best}).}
    \label{tab:accuracy_results}
    \begin{tabular}{lc}
        \toprule
        \textit{\textbf{Method}} & \textit{\textbf{Harmonic Mean}} \\ \midrule
        \textit{CADA-VAE~\cite{schonfeld2019generalized}} & $33.27\%$ \\
        \textit{f-CLSWGAN~\cite{xian2018feature}} & $30.18\%$ \\
        \textit{CVAE-ZSL~\cite{mishra2018generative}} & $31.74\%$ \\
        \textit{\tt \textbf{\modelname (Ours)}} & \accuracy \\
        \bottomrule
    \end{tabular}
    \vspace{-12pt}
\end{table}

\section{Conclusion, Limitations and Future Work}
In this work, we proposed a novel {\tt \modelname} architecture for generalized zero-shot learning of perceived emotions from 3D motion-captured gesture sequences. We used an adversarial loss to learn mappings between the gestures and the semantically-conditioned space of emotion words to classify gestures into both seen and unseen emotions. We evaluated our approach on the MPI Emotional Body Expressions Database (EBEDB), using feature-embeddings extracted from gestures and language-embeddings from \textit{word2vec}. Our proposed approach outperforms previous state-of-the-art algorithms for GZSL by $25$--$27\%$ 
on MPI EBEDB.

Our work has some limitations. Since \textit{word2vec} is a generic language-embedding model, not specific to emotions, it may not capture all aspects of psychological and emotional diversity. We, therefore, plan to affective-based semantics from words in the future. We also plan to incorporate more affective modalities, including speech and eye movements, to ensure a more robust classification. Furthermore, we plan to use the dimensional space of VAD (Valence-Arousal-Dominance) to learn relationships between disparate categorical emotions.

\appendix


\bibliography{aaai22}

\end{document}